\documentclass[sigconf]{acmart}
\usepackage{bbold}
\usepackage{bm}
\usepackage{multirow}
\usepackage{array}
\usepackage{caption}
\usepackage[export]{adjustbox}
\usepackage{titlesec}
\newcommand\oren[1]{\textcolor{blue}{OREN: {#1}}}

\usepackage{todonotes}
\usepackage{caption}
\presetkeys%
    {todonotes}%
    {inline,backgroundcolor=yellow}{}
\usepackage{verbatim}
\AtBeginDocument{%
  \providecommand\BibTeX{{%
    \normalfont B\kern-0.5em{\scshape i\kern-0.25em b}\kern-0.8em\TeX}}}

\setcopyright{acmcopyright}
\copyrightyear{2021}
\acmYear{2021}
\acmDOI{10.1145/1122445.1122456}

\acmConference[WSDM '21]{WSDM '21: ACM Web Search and Data Mining}{2021}{Jerusalem, Israel}
\acmBooktitle{ACM WSDM '21}
\acmPrice{15.00}
\acmISBN{978-1-4503-XXXX-X/18/06}
\begin{document}
\setlength{\abovedisplayskip}{3pt}
\setlength{\belowdisplayskip}{3pt}
% \captionsetup{belowskip=0pt}
% \addtolength{\parskip}{-1.5mm}
\setlength{\textfloatsep}{0.05cm}

% \title{Towards Robust Recommenders: Studying Visual Attacks}
\title{Towards Robust Recommenders: When Image Manipulation Overshadows Collaborative Filtering}
\title[A Black-Box Attack Model for Visually-Aware Recommenders]{A Black-Box Attack Model for Visually-Aware \\Recommender Systems}
\author{Rami Cohen}
\email{rami_cohen@intuit.com}
\affiliation{\institution{Intuit AI and Bar-Ilan University}}
\author{Oren Sar Shalom}
\email{oren.sarshalom@gmail.com}
\affiliation{\institution{Facebook}}
\author{Dietmar Jannach}
\email{dietmar.jannach@aau.at}
\affiliation{\institution{University of Klagenfurt}}
\author{Amihood Amir}
\email{amir@esc.biu.ac.il}
\affiliation{\institution{Bar-Ilan University}}

% Due to advances in deep learning, visually-aware recommender systems (RS) have recently attracted increased research interest. Such systems combine image information, represented as high-level feature vectors, with collaborative signals. Since item catalogs can be huge, recommendation service providers often rely on pre-trained image models such as ResNet and on images supplied by the item providers. As a result of this latter aspect, the image providers influence what is presented to users through the recommendations. In this work, we show that relying on such external sources can make an RS vulnerable to attacks, where the goal of the attacker is to unfairly promote certain \emph{pushed items}. Specifically, we demonstrate how a new \emph{visual attack} model can effectively influence the item scores and rankings in a black-box approach, i.e., without knowing the model parameters. Our main underlying idea is to systematically create small human-imperceptible perturbations of the pushed item image and to devise appropriate gradient approximation methods to incrementally raise the pushed item' score. Experimental evaluations on two datasets show that the novel attack model is effective even when the contribution of the visual features to the overall performance of the recommender system are modest.

\begin{abstract}
% Recommender systems (RS) help users mitigate information overload by suggesting relevant items in a personalized way.
% In recent years,
Due to the advances in deep learning, visually-aware recommender systems (RS) have recently attracted increased research interest. Such systems combine collaborative signals with images, usually represented as feature vectors outputted by pre-trained image models.
Since item catalogs can be huge, recommendation service providers often rely on
% pre-trained image models such as ResNet and on
images that are supplied by the item providers.
% , e.g., through an API
% . As a result,
% of this latter aspect
% the providers of the images influence what is presented to users through the recommendations.
In this work, we show that relying on such external sources can make an RS vulnerable to attacks, where the goal of the attacker is to unfairly promote certain \emph{pushed items}. Specifically, we demonstrate how a new \emph{visual attack} model can effectively influence the item scores and rankings in a black-box approach, i.e., without knowing the parameters of the model. The main underlying idea is to systematically create small human-imperceptible perturbations of the pushed item image and to devise appropriate gradient approximation methods to incrementally raise the pushed item's score. Experimental evaluations
on two datasets
show that the novel attack model is effective even when the contribution of the visual features to the overall performance of the recommender system
is modest.
\end{abstract}

 % Collaborative Filtering,

\maketitle

\section{Introduction}
% OLD Abstract
\begin{comment}
Recommender systems help users mitigate the information overload problem by suggesting relevant items. To generate personalized recommendations they apply machine learning algorithms, regretfully making them vulnerable to attacks that can be carried out by injecting malicious input. Thus far research on attacks focused on tampering with usage patterns, commonly known as shilling attacks. %A vendor might introduce fake users to the system, to increase the visibility of their own items in the recommender.
In this work we identify side-information as another possible input type that can be manipulated, focusing on images associated with the items.
We propose several models that differ by the amount of knowledge and capabilities required by the adversary. Using two benchmark datasets we prove the effectiveness of these approaches. We show that even when the visual features are of low contribution to the recommender system, they can be used by the adversary to completely alter the recommended lists.
%We present several trade-offs between the resources available to the adversary and the effect on the attacked recommender.
Our hope is this pioneering study will contribute to better understanding this new type of adversarial attacks and thus contribute to AI safety.
% We conclude this paper with a discussion on possible approaches to remedy the aforementioned attacks.
\end{comment}

During the last decades, with the increasing importance of the Web, Recommender Systems (RS) have gradually taken a more significant place in our lives. Such systems are used in various domains, from e-commerce to content consumption, where they provide value both for consumers and recommendation service providers. From a consumer perspective, RS for example help consumers deal with information overload. At the same time, recommender systems can create various types of economic value, e.g., in the form of increased sales or customer retention \cite{jannachjugovactmis2019,schafer1999recommender,fleder2009blockbuster, pathak2010empirical,jannach2019towards}.

The economic value associated RS makes them a natural target of \emph{attacks}. The goal of attackers usually is to \emph{push} certain items from which they have an economic benefit, assuming that items that are ranked higher in the recommendation list are seen or purchased more often. Various types of attack models for RS have been proposed in the literature \cite{si2020shilling, gunes2014shilling}. The large majority of these attack models is based on injecting fake profiles into the RS. These profiles are designed in a way that they are able to mislead the RS %underlying machine learning models
to recommend certain items with a higher probability.

Nowadays, many deployed systems are however not purely collaborative anymore and do not base their recommendations solely on user preference profiles. Instead, oftentimes hybrid approaches are used that combine collaborative signals (e.g., explicit ratings or implicit feedback) with side information about the items.
% Historically, the used side information related to item meta-data or the \emph{content}, in case of the recommendation of textual objects like news or Web pages.
In the most recent years, it was moreover found that item images, i.e., the visual appearance of items, can represent an important piece of information that can be leveraged in the recommendation process. These developments led to the class of \emph{visually-aware} recommender systems. Fueled by the advances in deep learning, a myriad of such approaches were proposed in recent years \cite{he2016vbpr, bostandjiev2012tasteweights,kim2004viscors,linaza2011image, bruns2015should,iwata2011fashion,deldjoo2016content,he2016sherlock}.

In many domains, catalog sizes can be in the millions. Hence, from a practical point of view, it can be challenging or impractical for recommendation service providers, e.g., an e-commerce shop or media streaming site, to capture and maintain item images (e.g., product photos) for the entire catalog by themselves.  Therefore, a common practice is to rely on the \emph{item providers} to make  their own images available to recommendation service providers by a dedicated API. This is a cross-domain practice, where e-commerce companies like Amazon\footnote{\url{https://vendorcentral.amazon.com/}}
% or eBay\footnote{\url{http://developer.ebay.com/DevZone/XML/docs/Reference/eBay/UploadSiteHostedPictures.html}}
enable sellers to upload photos and content platform like YouTube\footnote{\url{https://support.google.com/youtube/answer/72431}} allow creators to upload their thumbnails. With the term \emph{item providers}, we here refer to those entities (individuals or organizations) who act as suppliers of the items that are eventually recommended and who benefit economically from their items being listed in recommendations.

In modern visually-aware recommender systems, the features of the uploaded item photos, as discussed above, can influence the ranking of the items. Therefore, the item provider might be an \emph{attacker} and try to manipulate the images in a way that the underlying machine learning model ranks the item to be pushed higher in the recommendation lists of users. We call this novel way of manipulating a recommender system a \emph{visual attack}.

While the literature on adversarial examples is rich, only few works exist on their use in the context of RS \cite{di2020taamr,tang2019adversarial}. Moreover, these works rely on impractical assumptions and require the attacker to have access to the internal representation of the RS. In contrast, the proposed algorithm assumes milder and more realistic assumptions.
% While the literature on adversarial examples in deep learning is rich, limited works exist yet on their use in the context of RS. Notable recent works include \cite{di2020taamr} and \cite{tang2019adversarial}. However, these works follow a ``white-box'' approach and assume that an attacker can either manipulate the \emph{internal representation} of the images as used by the RS, or has complete knowledge about the system. Since these assumptions may be strong in practice, we present in this paper a novel ``black-box'' visual attack model.
The main idea of the approach is to systematically manipulate the provided item images in a human-imperceptible way so that the algorithm used by the recommender predicts a higher relevance score for the item itself. Technically, we accomplish this by devising appropriate gradient approximation techniques. The contributions of our work are as follows:
\begin{itemize}
  \item We identify a new vulnerability of RS, which emerges in situations where item providers are in control of the images that are used in visually-aware recommenders;
  \item We propose novel technical attack models that exploit this vulnerability under different levels of resources available to the adversary;
  \item We evaluate these models on two datasets, showing that they are effective even in cases where the visual component only plays a minor role for the ranking process;
  \end{itemize}
To make our results reproducible, we share all data and code used in our experiments online\footnote{https://github.com/vis-rs-attack/code}.

\section{Background and Related Work}\label{sec:relared}
%We now review three prerequisite research areas. %previous
%the architecture of
% visually-aware RS, attack models, and
%the use of
% adversarial examples in image classification.
We review three prerequisite research areas: visually-aware RS, attack models, and the use of adversarial examples in image classification.

\subsection{Architecture of Visually-Aware RS}\label{ssec:vrec}
% The visual appearance of the recommendable items has been considered in a variety of recent approaches to build visually-aware recommender systems, as mentioned above \cite{he2016vbpr, bostandjiev2012tasteweights, kim2004viscors, linaza2011image, bruns2015should, iwata2011fashion, deldjoo2016content, he2016sherlock}. Typically, these systems combine collaborative information with image information in a hybrid approach.

With respect to their technical architecture, state-of-the-art visually-aware recommender systems commonly incorporate high-level features of item images that were previously extracted from fixed pre-trained deep models. The system therefore invokes a pre-trained Image Classification model (IC), e.g., ResNet \cite{he2016deep} or VGG \cite{simonyan2014very}, which returns the output of the penultimate layer of the model, dubbed the \emph{feature vector} of the image. The advantage of using such pre-trained IC model is that they were trained on millions of images, allowing them to extract feature vectors that hold essential information regarding the visualization of the image.

\begin{figure}[t!]
%\vspace{-0.2cm}
% \captionsetup{belowskip=0pt}
\setlength{\belowcaptionskip}{-10pt}
    % \centering
%         \fbox{\includegraphics[clip, trim=2.5cm 6cm 9cm 4cm, width=0.6\textwidth]{architecture.pdf}}
        \includegraphics[scale=0.5]{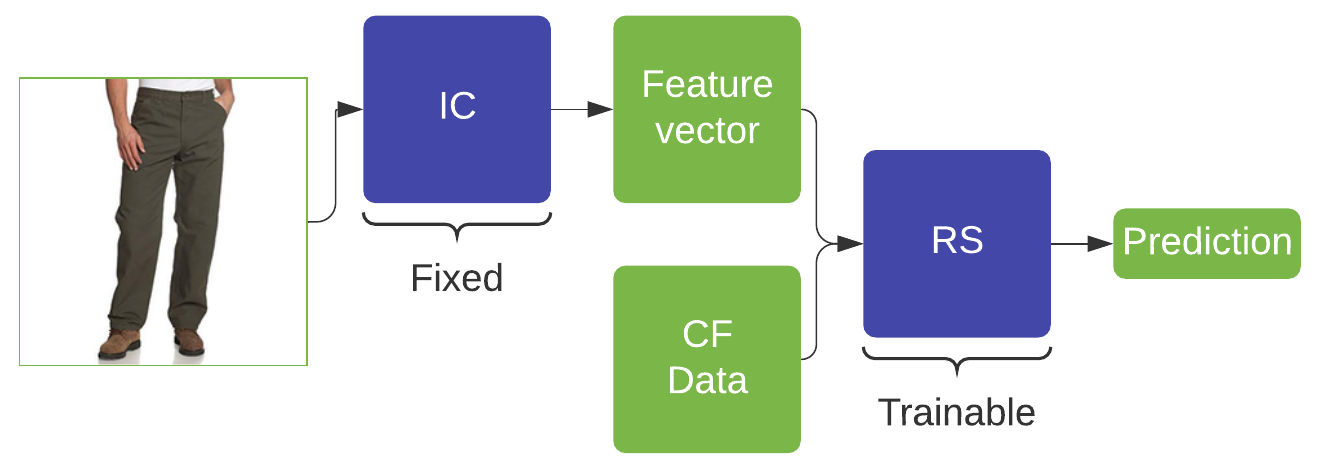}
\caption{Common Architecture of a Visually-aware RS}
\label{fig:vrs}
\vspace{0.4cm}
\end{figure}

%As a result,
However, the feature vector is \emph{not} optimized for generating relevant
recommendations. It is thus left to the recommender to learn a transformation layer
that extracts valuable information with respect to recommendations. Usually this transformation is accomplished by multiplying the feature vector by a learned matrix $E$. Since the image is passed through a deterministic transformation, it defines a revised input layer having the feature vector replacing the pixels.
A generic schematic visualization of this approach is depicted in Figure \ref{fig:vrs}.

% \squeezeup
VBPR \cite{he2016vbpr} was probably the first work to follow this approach and extends BPR \cite{rendle2012bpr} to incorporate images. It maintains two embedding spaces, derived from the usage patterns and the item visualization. The final predicted score results from the sum of the scores of these two spaces. Sherlock \cite{he2016sherlock} adds the ability to model the feature vector with respect to the item category, by learning a dedicated transformation matrix $E_i$ per each category $i$.
Later, DeepStyle \cite{liu2017deepstyle} improved the category-aware approach by learning a single embedding matrix which maps the categories to the latent space of the images. Thus, a single transformation matrix $E$ is sufficient. The authors of this paper showed that the reduced number of parameters leads to enhanced performance.

% Generally, images represent an integral part of the modeling process in these and in similar approaches. An attacker might therefore try to tamper with the images of the items to be pushed. Specifically, the tampered image should increase the probability that the items are recommending to target users while remaining indistinguishable from the genuine image. In our work, we demonstrate how such attacks can be effective even with very limited knowledge.
% %To the best of our knowledge, this is the first work that manipulates the pixel intensities of images in order to enable push attacks.
% DJX: I think at least two other methods mentioned above tried that, although with different assumptions and goals.
% \vspace{-0.15cm}
\subsection{Attacks on Recommender Systems}
%\subsection{Effectiveness of Attacks on RS}
The best explored attack models in the literature are based on the injection of fake users into the system, called \emph{shilling}. These fake users leave carefully designed feedback that is intended to both seem genuine and to disrupt the output of the RS  such that the pushed items will be recommended to the target users \cite{si2020shilling,gunes2014shilling}. This shilling attack can be highly effective for cold items, a ubiquitous phenomenon in real-life datasets \cite{sar2015data}.

The effectiveness of an attack is usually measured by the amount of target users that get the pushed item as a recommendation. In the case of shilling attacks, there is a positive correlation between effectiveness and adversary resources, where resources are usually composed of knowledge and capacity. The relevant knowledge includes several aspects, for example, knowing the type of the deployed model, %(e.g., whether the engine is k-nearest neighbors or matrix factorization),
observing the values of the parameter set, or even having read access to the rating dataset that empowers the model. The capacity refers to the amount of shilling users and fake feedback the adversary can inject. Trivially,
% as the adversary possesses more knowledge and has greater capacity,
with more knowledge and capacity
the effectiveness of the attacks increases \cite{mobasher2007attacks}. As we will show, the trade-off between resources and effectiveness also exists in visual attacks. However, the relevant resources in these types of attacks are entirely different from shilling attacks.

There are only very few existing works that rely on image information to attack an RS, as mentioned above, \cite{di2020taamr} and \cite{tang2019adversarial}.
However, these works assume that the attacker can either manipulate the \emph{internal representation} of the images as used by the RS, or has complete knowledge about the system.
% Both assume either extended knowledge about the system or even control over the internal image representation.
Moreover, in the case of \cite{di2020taamr}, the goal is to change the predicted category classification of an item, whereas in our approach we aim to push items to the top-k lists of target users.

\subsection{Adversarial Examples in Image Classification}
\label{subsec:adversarial-examples}
Adversarial examples in machine learning are inputs that seem almost identical to examples from the true distribution
of data that the algorithm works on. However, they are designed to deceive the model and cause it to output incorrect
predictions. In the context of image classification, adversarial examples are constructed by adding imperceptible changes
to genuine images, with the aim of crossing the decision boundary of the classifier (e.g., \cite{szegedy2013intriguing}).
Adversarial examples can be generated either through \emph{white-box} (WB) or \emph{black-box} (BB) approaches.

\paragraph{White-box:} This attack assumes the adversary has complete knowledge of the model and can compute gradients using backpropagation. Unlike regular stochastic gradient descent (SGD), which updates the model \emph{parameters} in the \emph{opposite} direction of the gradient, the adversary updates the \emph{input} in the direction of the gradient. Let $\eta$ be the adversarial direction. Then an adversarial example is generated as $\Tilde{x}=x + \epsilon\eta$.
% a predefined threshold is set to keep the adversarial change undistinguished to the human eye.
%Many approaches have been proposed to control the degree of visual dissimilarity. Examples include the ``fast gradient sign method'' \cite{goodfellow2014explaining}, optimized for $L_\infty$, which approximates this search problem by a binarization of the gradient, computed as $\eta\leftarrow sign(\eta)$. Early stopping was suggested by \cite{kurakin2016adversarial} and Madry et al. \cite{madry2017towards} used the ``projected gradient descent'', to limit the $\ell_2$-norm.

\paragraph{Black-box:}
Such an attack requires a milder assumption: the ability to query the model \cite{chakraborty2018adversarial}. In the absence of gradients to guide the adversary, it invokes the model multiple times with various perturbations within a small volume around a target image and uses the associated feedback to infer the right direction. However, due to the high-dimensionality of the \emph{pixel space}, even efficient attacks need at least tens of thousands of requests to the classifier, limiting their applicability \cite{ilyas2018black,cheng2018query,tu2019autozoom, guo2019simple}.
% Black-box attacks can either aim to manipulate a system's predicted scores or at the resulting rankings.
% In our work, we propose attack models both for the goals of score and ranking manipulations. Furthermore,
Our proposed approach does not operate on the pixel space, which in turn limits the number of queries to the model.
%The black-box threat is further subdivided according to the specification of the attacked model, which may return the \emph{score} of each predicted category, or only its \emph{ranking}. Once again, there is a trade-off between the effectiveness of the attack and the specification level of the model.

% To keep the adversarial changes imperceivable, One goal here is that
To assure the updated image does not deviate too much from the original one, some metric $d(\Tilde{x}, x)$ is usually defined to quantify the visual dissimilarity attributed to the adversarial change. Commonly, $d$ is computed by either $\ell_2$ or $\ell_\infty$, and the goal is to deceive the model while keeping a low dissimilarity score to keep the change undistinguishable by humans.
Various approaches were proposed to control the degree of visual dissimilarity. Examples include the ``fast gradient sign method'' \cite{goodfellow2014explaining}, optimized for $L_\infty$, which approximates this search problem by a binarization of the gradient, computed as $\eta\leftarrow sign(\eta)$. Early stopping was suggested by \cite{kurakin2016adversarial} and Madry et al. \cite{madry2017towards} used the ``projected gradient descent'' to limit the $\ell_2$-norm.

Both white-box and black-box attacks can improve effectiveness by working iteratively and taking multiple small \emph{steps} in the direction of the gradient. For instance, the ``Iterative Gradient Sign'' technique \cite{kurakin2016adversarialb} applies multiple steps of the ``fast gradient sign method'' \cite{goodfellow2014explaining} and reports boost in performance.

\section{Proposed Technical Approach}
\label{sec:technical-approach}
In this section we demonstrate how an attacker can exploit the identified vulnerability in visually-aware RS that are based on an architecture as sketched in Figure \ref{fig:vrs}. We will first elaborate how a white-box attack can be mounted and then move on to more realistic black-box attacks. In particular we will show how to attack both predicted scores and rankings.

\subsection{Preliminaries and Assumptions}
The proposed visual attacks solely modify the visualizations of the pushed items. Therefore, pushing a certain item would not change the score of any other item. As a corollary, gradient computations in attacks with several pushed items are independent of each other. Hence, in this paper we assume there is a single pushed item.

There are three types of target users: specific users, segments or general population. Focusing on specific users is beneficial as it is equivalent to personal advertising and may exploit the influence of opinion leaders. However, in order to attack a specific user, the adversary has to know the consumption history of that user. Then, that user can be cloned by introducing a new user to the RS with the exact same usage patterns. The attack is carried out on the cloned user and the target user will, as a result, be affected by the attack.

%We note that in many practical scenarios, such information is practically unattainable and we demonstrate this attack mainly as a proof of concept. Furthermore, since the impact of attacking a certain user is limited, potential adversaries would probably prefer to focus on other types of target users.

Approaching general population dispenses with the need to have prior knowledge about individual users, but it may result in huge exposure of the pushed items to irrelevant users. %Increasing the visibility of the target item to irrelevant users is unlikely to impact the promotional activity and even might lead to reputational risk.
A segmented attack is an in-between approach, where the adversary targets a set of users with similar tastes.
As we will explain later, even in the general population and segmented attacks we can safely assume that the adversary attacks a \emph{single} mock user that was \emph{injected by the attacker} to the system.

% The adversary can choose a particular market segment that is likely to find that item relevant.
% A producer wishing to promote the target item would probably want to increase the frequency it is recommended to a mass of people, and not just to specific users. However,

\subsection{An Upper Bound for Visual Dissimilarity}
In the presented attack models, the item images are systematically perturbed in a way that the related item's score or ranking is improved. This process is done step-wise and at each step more noise is added to the image. In order to prevent the attack from being detected and to ensure good image quality, it is important to limit the number of steps.

To assess the perceptual visual distortion caused by the adversarial changes, we use the widely adopted \emph{objective measure} Structural SIMilarity (SSIM) index \cite{wang2004image}. This index considers image degradation as perceived change in structural information. The idea behind structural information is that pixels have stronger dependencies when they are spatially close.
The SSIM index also incorporates, with equal importance, luminance masking and contrast masking terms. These terms reflect the phenomena where image distortions are more visible in bright areas and in the background, respectively.
%To compute the SSIM index between two images, local SSIM indices are computed over small corresponding windows and their average is returned. Formally, Let $x$ and $y$ be windows of a clean image and of an adversarial perturbation of it. The SSIM index is computed as:
%$S(x, y)=\frac{(2\mu_x\mu_y + C_1)(2\sigma_{xy} + C_2)}{(\mu_x^2 + \mu_y^2 + C_1)(\sigma_x^2 + \sigma_y^2 + C_2)}$
%where $\mu$, $\sigma^2$ and $\sigma_{xy}$ stand for mean, variance and covariance, respectively, and $C_1,C_2$ are small constants to %avoid division by zero. %The index ranges between -1 and 1, and value 1 is obtainable only for identical images.
%We computed the index values using the scikit-image library\footnote{\url{https://scikit-image.org/docs/dev/auto_examples/transform/plot_ssim.html}}.

% Figure \ref{fig:vis_steps} illustrates the aggregate changes along with their associated SSIM scores after a number of update steps on a random image. As can be seen, until about 20 steps the changes are fairly imperceptible. Afterwards the attack might be revealed as some noise is noticed mostly in the background, which shows some unnatural texture. This phenomenon repeats in the dataset, as
The average SSIM indices after 10, 20 and 30 steps are: 0.965, 0.942 and 0.914, respectively (higher is better), which entails a noticeable degradation after 20 steps. Most images have a solid white background, however beyond 20 steps the attack might be revealed as some noise is noticed mostly in the background, which shows some unnatural texture.
Therefore we set the maximal allowed number of steps to 20 in our experiments.

% \begin{figure*}
% \includegraphics[scale=0.6]{vis-steps-ssim-1.pdf}
% %\caption{Illustration of the visual change in images with increasing number of update steps}
% \caption{Illustration of visual changes with an increasing number of update steps; notice the background texture.}
% \label{fig:vis_steps}
% %\vspace{-0.5cm}
% \end{figure*}

% \vspace{-0.1cm}
\subsection{A White-Box Attack}\label{sec:wb_attack}
To define an upper bound for the effectiveness of visual attacks as investigated in our paper, we show how a white-box attack can be designed. This attack assumes the adversary has a \emph{read} access to the parameters of the model, which is in general not realistic.

At inference time, to allow personalized ranking, the score of user $u$ for item $i$ is computed by a visually-aware recommender as $s_{u,i}=f(u, i, p_i)$, where $p_i$ is the image associated with item $i$. Note that the only assumption we make on $f(\cdot)$ is that it is differentiable. To avoid clutter, we omit the subscripts whenever their existence is clear from the context. To manipulate the image, the adversary computes the partial derivatives $\frac{\partial s}{\partial p}$ and updates the pixels in that direction, while keeping the distorted image as close as possible to the genuine one. Following \cite{goodfellow2014explaining}, we binarize these partial derivatives by taking their sign, as our preliminary experiments asserted it leads to faster convergence. We note that any future improvements in adversarial examples for images may contribute to the effectiveness of this proposed approach.

The visual attack, as mentioned, is performed in small \emph{steps}. Each step $t+1$ aims to increase the predicted score of the pushed item by updating the image as follows:
\begin{equation}\label{eq:step}
p^{t+1}=p^t + \epsilon\cdot sign(\frac{\partial s}{\partial p^t})
\end{equation}
where $p^0$ is the genuine image and $\epsilon$ is a small constant that controls the step size so the adversarial changes will be overlooked by the human eye. 

%This type of attack mainly serves as an upper bound for attack effectiveness since assuming the adversary has access to the parameter set of the model might be unrealistic. Therefore, potential attackers need to resort to black-box approaches, detailed in the next section.

\subsection{A Black-Box Attack on Scores}
As in the white-box attack, the goal of the adversary is to compute the gradient of the score $s$ with respect to all pixels in the image $p$ in the corresponding image, i.e., $\frac{\partial s}{\partial p}$. However, as explained in Section \ref{subsec:adversarial-examples}, black-box attacks in image classification usually require an enormous amount of requests to the classifier because they work in the \emph{pixel space}. Applying similar techniques in visual attacks would therefore be impractical.

Remember that the RS subnetwork is unknown to the attacker (neither its parameters or even its architecture).
However, remember that in our target architecture for visually-aware RS (Figure \ref{fig:vrs}), we assume that the system leverages
%the capabilities of
a pretrained IC model. As discussed in Section \ref{ssec:vrec}, such IC models are publicly available and held constant during training, while the rest of the network is optimized for the specific task and data.  In our case, given a new image, the RS extracts a feature vector $f_i$ from image $p_i$ by feed forwarding it through the IC and outputting the penultimate layer.

This situation now allows us to apply the chain rule and the multiplication between the partial derivatives of the score $s$ with respect to $f$ and the Jacobian of $f$ with respect to $p$ can be computed. Formally:
$\frac{\partial s}{\partial p}=\frac{\partial s}{\partial f} \cdot \frac{\partial f}{\partial p}$. Upon computing this value, a step is made as defined in Eq. \ref{eq:step}.
Since we assume the parameter set of the image processing unit is known, $\frac{\partial f}{\partial p}$ can be analytically computed using backpropagation.

% \vspace{-0.3cm}
\subsection{Computation of $\frac{\partial s}{\partial f}$}
Decoupling the partial derivative into two quantities allows to avert numerical computations in the pixel space. However, the computation of the partial derivatives $\frac{\partial s}{\partial f}$ under the black-box model of the RS is rather complicated because $f$ is not the actual input layer.

To illustrate the challenge, consider the following approach for numerical computation of the partial derivatives, where $d$ denotes the dimensionality of $f$. Create $d$ perturbations of $p$, such that each perturbation $p^k$, for $0\leq k < d$, under the constraint $\left\| p-p^k \right\| \leq \delta$, yields a feature vector $f^k=f + \epsilon\cdot \mathcal{I}(k)$ for some predefined $\delta$ and arbitrary small $\epsilon$, where $\mathcal{I}(k)$ is the one-hot vector with $1$ for $k$-th coordinate and $0$ elsewhere. That is, each perturbation isolates a different dimension in $f$. Then, replace the genuine image $d$ times, each time with a different perturbation, and obtain from the RS a new personalized score $s^k$. Now the numerical partial derivative $\frac{\partial s}{\partial f^k}$ is given by $\frac{s^k - s}{\epsilon}$. However, since each pixel in $p$ holds a discrete value and affects all dimensions in $f$, finding such a set of perturbed images is intractable.

Furthermore, in the absence of the ability to directly control $f$, it is not clear how to apply existing methods of gradient estimation. For instance, the widely used NES \cite{wierstra2008natural} requires to sample perturbations from a Gaussian distribution centered around the input $f$; even the seminal SPSA algorithm \cite{spall1992multivariate} demands a \emph{symmetric} set of perturbations.
As such methods are unattainable, we devise a novel approach to estimate the gradients. This method does not have any constraints on the structure of the perturbations. Moreover, the number of perturbations required by this method is \emph{sublinear} in $d$.
To this end, we first present a method that requires exactly $d$ perturbations and then show how this amount can be reduced.
The attacker creates $d$ random perturbations of the image uniformly drawn from the $L_\infty$-ball centered around $p$ with radius $\delta$, for some arbitrary small $\delta$, and obtains their corresponding feature vectors $f^k$ together with the associated personalized scores $s^k$.
Zou et al.~\cite{zou2019lipschitz} proved that CNN-based architectures are Lipschitz continuous, and empirically showed that their Lipschitz constant is bounded by $1$.
Consequently, since each perturbation is similar to the original image, up to a tiny difference $\delta$, also the deltas in the feature vectors $\left\| f^k-f \right\|$ are expected to be epsilonic, making them suitable for numeric computation of $\frac{\partial s}{\partial f}$.
Using matrix notation, the set of vectors $\{f^k\}$ and scalars $\{s^k\}$ are referred to as matrix $\bm{F}\in \mathbb{R}^{d\times d}$ and vector $\bm{s}\in \mathbb{R}^d$, correspondingly.

This translates into a linear system $\bm{F}x=\bm{s}$ such that its solution $x$ is the numerical computation of $\frac{\partial s}{\partial f}$. Keep in mind that a linear system is shift invariant. That is, the same solution $x$ is obtained upon subtracting $f^T$ from each row of $\bm{F}$ and subtracting $s$ from each component in $\bm{s}$. Now the coefficient matrix is the change in $f$ and the dependent variable is the resulting change in score:
\begin{equation}\label{eq:derivative}
    x = (\bm{F}-f^T)^{-1}(\bm{s}-s) = \Delta\bm{F}^{-1}\Delta\bm{s}=\frac{\partial s}{\partial f}
\end{equation}

Note that a single solution exists if and only if $\bm{F}$ is nonsingular. Since it is obtained by random perturbations, it is almost surely the case \cite{kahn1995probability}; otherwise a new set of perturbations can be drawn.

\paragraph{Reducing the Number of Perturbations}
%we assume a negligible overhead caused by uploading a single image to the server. However, the rank-feedback attack requires to upload $d$ perturbations of the image per each step.
Each perturbation entails uploading a new image to the server. In ResNet, for example, $d=2,048$, which results in a large amount of updates. We next address this problem and suggest a trade-off between effectiveness of each step and amount of required image updates.

Note that although Eq.~\ref{eq:derivative} requires $d$ perturbations for finding a single solution, an approximation can be found with fewer perturbations. Formally, let $d'$ be the number of perturbations the adversary generates, for some $d'<d$. This defines an underdetermined system of linear equations, and among the infinite solutions as the estimated gradients, the one with the minimal $\ell_2$ norm can be found \cite{cline1976l_2}.
Another advantage of reducing the number of perturbations is the ability to generalize. There is an unlimited number of possible perturbations, and every subset of $d$ perturbations defines another linear system, with another exact solution. Those exact solutions vary in their distances from the true, unobserved derivatives. However, since the adversary randomly select a single subset of perturbations, the exact solution may ``overfit'' to this instantiation of the linear system, and not generalize to other solutions that could be yielded from other subsets.
Such behavior might be highly problematic, and must be addressed correspondingly. To remedy this problem, regularization comes in useful. Reducing the number of perturbations, and finding a solution with minimal $\ell_2$ norm can be thought of as a means of regularization, and can potentially improve performance. Indeed, as we show in Section \ref{sec:experiments}, our approach not only reduces the required amount of perturbations also leads to improved performance.

% Going back to the overall attack scheme, remember that to perform the attack after each step of approximating gradients, an actual image has to be created and to be uploaded to the RS.

% As a result each next step will be computed based on the updated one. This procedure has two consequences:
% \begin{itemize}
%     \item The values at $p_{t+1}$ are float numbers, which should be rounded to integers in order to make an actual image. Adversarial examples for images exploit delicate nuances, and rounding all pixels after each step might diminish them. This is yet another reason to limit the number of steps.
%     \item Updating the corresponding image might consume some time. Since it does not require to retrain the model, and it is common to have an API for item producers to update their images, its overhead is bearable.
% \end{itemize}

\subsection{A Black-Box Attack on Rankings}
The attack described in the previous section is suited to target RS that reveal the scores, e.g., in the form of predicted ratings. A more common situation, however, is that we can only observe the personalized rankings of the items. Here, we therefore describe a novel black-box method that operates on the basis of such rankings.

In this attack, the adversary generates $d$ perturbations as well, and the scores $s_k$ should be recovered by using only their ranking $r_k$. If the score distribution of the $d$ perturbations was known, the inverse of the CDF could accurately recover the scores. A common assumption is that item scores follow normal distribution \cite{steck2015gaussian}. However, even if it is true in practice, its mean and variance remain unknown. As observed from Eq. \ref{eq:derivative}, the solution is shift invariant, so the mean of the distribution is irrelevant, but the variance does play a role.

\paragraph{Proposed Approach}
To bypass this problem we note that perhaps the entire catalog's scores follow a Gaussian distribution, where various items are involved. However, the differences in the perturbations' scores stem only from small changes to the image, while the pushed item itself is fixed. Therefore, we assume the scores reside within a very small segment $[s_{min},s_{max}]$ of the entire distribution. If the segment is small enough, it can be approximated by a uniform distribution. Hence, a mapping between ranking and score is given by $s_k=\frac{N-r_k}{N}(s_{max}-s_{min})$, where $N$ is the number of items in the catalog.

The remaining question is how to compute these values without knowledge of $s_{min}$ or $s_{max}$. First, as Eq. \ref{eq:derivative} is shift invariant, we can assume $s_{min}=0$. Second, because we solve a linear system, multiplying the dependent variable (scores) by a scalar, also scales the solution (derivatives) by the same factor. Since we take the sign of the gradient, scaling has no effect on the outcome. Thereby, we can assume $s_{max}=1$, and therefore $s_k=1-\frac{r_k}{N}$.

Remarkably, this simple approach performed well in our initial experiments. In these experiments, we created a \emph{privileged baseline} that receives the empirical standard deviation as input (information that is not accessible to the adversary), which allows it to recover the scores using the inverse of the CDF of the appropriate normal distribution.
% computes the empirical standard deviation by observing the actual scores (information that is not accessible to the adversary).
Nevertheless, assuming uniform distribution outperformed that baseline. We explain this phenomenon by the fact that the perturbations do not perfectly follow a Gaussian distribution, which leads to inconsistency in the inferred scores.

%\paragraph{Partial Ranked List}
\paragraph{Dealing with Partial Ranking Knowledge:}
One possible limitation of our algorithm is the assumption that the complete ranking of all items in the catalog is revealed by the RS.
However, in some cases not all catalog items are shown even when the user scrolls down the recommendation list.
%only the top items appear, and even scrolling down the recommendation list is limited to a small fraction of the catalog.
%Therefore, there is a negligible probability that the ranking of the pushed item is shown to the attacker, which averts the
Therefore, the pushed item might not be shown to the attacker, which averts the proposed attack since no gradients could be computed.
To overcome this obstacle, a surrogate user $u'$ is introduced by the attacker and set as the attacked one. $u'$ maintains two properties: 1) receives the pushed item in the top of the list, so gradients can be computed; 2) is similar to $u$, so the computed gradients are useful for the attacker.

Specifically, $u'$ is identical to the target user $u$, but has an added item $i'$ in the consumption history. This item is carefully selected, such that it makes the pushed item penetrate to the top items of user $u'$, thus allowing to compute $\frac{\partial s_{u',i}}{\partial f_i}$. Note that this partial derivative is close to $\frac{\partial s_{u,i}}{\partial f_i}$ but is not identical, as it is affected by $i'$.
Let $n$ denote the number of items in the history of $u$ and $u(i)$ a user with only item $i$ in the consumption history. Naturally the relative contribution of $i'$ to the gradient decreases with $n$.
As the attacker has no prior knowledge on the underlying RS, by simply assuming a linear correlation, the contribution of $i'$ is counteracted as follows: $\frac{\partial s_{u,i}}{\partial f_i} = \frac{(n+1)\cdot \frac{\partial s_{u',i}}{\partial f_i} - \frac{\partial s_{u(i'),i}}{\partial f_i}}{n}$. Once $i$ appears in the top of $u$'s recommendations, the attack is performed directly on $u$.
Item $i'$ is selected from the top recommendations for $u(i)$. In most cases, adding item $i'$ to the consumption history of $u$ is not enough to place $i$ in the top of the list of $u'$. Therefore, several attack steps are performed on $u(i')$, to increase $s_{u(i'),i}$ until $i$ appears at the top of the list of $u'$. Overall, in order to push item $i$, the algorithm requires to create 2 additional users: $u(i)$ and $u(i')$, which is an attainable request in general.

\section{Experimental Evaluation} \label{sec:experiments}
%The purpose of this section is to empirically prove the destructive potential of adversarial examples in recommenders.
We demonstrate the effectiveness of the attack models by conducting experiments on several combinations of RSs and datasets.
% To assess the effectiveness of the proposed attack models we conducted experiments on several combinations of visually-aware recommenders and datasets. %, and report the effectiveness of each type of adversary.

\subsection{Experiment Setup}
\subsubsection{Datasets}
%We demonstrate the effect of visual attacks on
We relied on two real-world datasets that were derived from the Amazon product data \cite{mcauley2015image}.
The first dataset is the Clothing, Shoes and Jewelry dataset (named ``Clothing'' for short) and the second one is the Electronics dataset.
These datasets differ in all major properties, like domain, sparsity and size, which shows visual attacks are a ubiquitous problem. Arguably chief among their distinctive traits is the importance of visual features, quantitatively approximated by the gain in performance attributed to the ability to model images.
To compute this trait, dubbed as ``visual gain'', we use AUC \cite{rendle2012bpr} as a proxy for performance. Since VBPR \cite{he2016vbpr} extends BPR \cite{rendle2012bpr} to model images, the visual gain is computed as the relative increase in the AUC obtained by VBPR over BPR.
% We dub this trait ``visual gain'', computed as the ratio of AUCs obtained by VBPR\cite{he2016vbpr} and BPR\cite{rendle2012bpr}.
% excess performance of VBPR\cite{he2016vbpr} comparing to BPR\cite{rendle2012bpr}.

Table \ref{tab:datasets} summarizes statistics of these datasets.
It is noticed that the visual gain in the electronics dataset ($2\%=\frac{0.85}{0.83}-1$) is much lower than in the clothing dataset ($23\%=\frac{0.79}{0.64}-1$).
We therefore expect that visual features in the electronics domain would be of lesser importance, and as a consequence the RS will only moderately consider them.
Nevertheless, we show that visual attacks are effective even in this domain.
% Thus, it might be very challenging to carry out a visual attack in this domain.

\begin{table}[t!]%[htb!]
\centering
 \begin{tabular}{|l|c|c|c|c|}
 \hline
 Dataset & \#users & \#items & \#feedback & Visual gain \\
 \hline %\hline
 Clothing & 127,055 & 455,412 & 1,042,097 & 23\%  \\
 Electronics & 176,607 & 224,852 & 1,551,960 & 2\%  \\
 \hline
 \end{tabular}
\caption{Dataset statistics after preprocessing}
\label{tab:datasets}
% \vspace{-0.4cm}
\end{table}

\subsubsection{Attacked Models}
We experiment with two visually-aware RS as the underlying attacked models.
The first is the seminal algorithm VBPR \cite{he2016vbpr}.
To assess the effect of visual attacks in a common setting where \emph{structured} side information is available, the second model is DeepStyle \cite{liu2017deepstyle}, which also considers the item categories.
We train a model for each combination of dataset and underlying RS, yielding 4 models in total. We randomly split each dataset into training/validation sets, in order to perform hyperparameter tuning. Both algorithms are very competitive and the obtained AUC of VBPR (DeepStyle) on the validation set of the Clothing dataset is 0.79 (0.80) and 0.85 (0.86) in the Electronics domain.

\subsubsection{Visual Features}
The underlying image classification model is ResNet \cite{he2016deep}. This is a 50 layer model, trained on 1.2 million ImageNet \cite{deng2009imagenet} images. For each item to be modeled by the RS, its associated image is first fed into the IC model and the penultimate layer of size 2,048 is extracted as the feature vector. % Figure \ref{fig:vrs} schematically depicts this process.

\subsubsection{Evaluation Metric}
The purpose of attacks is to increase the visibility of the pushed items in top-k recommendations. Therefore, we follow \cite{sarwar2001item,mehta2008attack} and evaluate the effectiveness of an attack using the Hit-Ratio (HR), which measures the fraction of affected users by the attack. Formally, let $\mathcal{S}$ be the set of user-item pairs subject to an attack. For each $(u,i)\in\mathcal{S}$, the Boolean indicator $H^k_{u,i}$ is true if and only if item $i$ is in the top-k recommendation to user $u$. Then the Hit Ratio is defined as follows:
$$HR@k = \frac{1}{|\mathcal{S}|}\sum_{(u,i)\in\mathcal{S}} H^k_{u,i}$$

In real-life datasets with many items, the probability of a random pushed item to appear in the top of the recommendation list is extremely low. Therefore HR directly measures the impact of the attack, as the number of pre-attack hits is negligible or even zero.

%\subsection{Methodology}
%\subsection{Target Populations and Baselines}
%We conducted experiments using different target populations and we furthermore included an easy-to-detect baseline in our experiments to gauge the relative performance of our attacks.

\subsection{Varying the Target Population}
We design dedicated experiments for each type of target populations to reflect the different assumed capabilities of the adversary.

\subsubsection{Specific-users attack}
In this threat, the adversary knows the consumption list of the attacked user and hence can impersonate that user by introducing to the RS a new user with the same list.
To simulate these attacks, we randomly select 100 user-item pairs to serve as the attacked users and pushed items.

\subsubsection{Segmented attack}
Following \cite{burke2005segment}, a segment is a group of users with similar tastes, approximated by the set of users with a common specific item in their interaction histories. For instance, a segment can be defined as all users who purchased a certain Adidas shoe. We name the item that defines the segment as the \emph{segment item}\footnote{Supporting multiple segment items can be trivially implemented as well.}.
To perform this attack, the adversary creates a single mock user who serves as the target user, whose interaction history includes only the segment item.
We do not assume that the adversary possesses information about usage patterns in general or knows which item defines the optimal segment item. Therefore, the segment item is simply chosen by random among all items in the same category of the pushed item.
In our experiments we chose 100 random pairs of pushed and target items. We emphasize that the actual target users are those who interacted with the segment item, but we do not assume the adversary knows their identities.
% Furthermore, we do not assume the attacker knows which item defines the optimal segment for the pushed item, and therefore we choose it by random.
We report the effectiveness of the attack on the actual set of target users, which is obscured from the adversary.
% As an example, if the attacker is a producer of sport shoes, then the concentrated set of users may be those who purchased a certain Adidas shoe.

\subsubsection{General population}
This attack assumes that the adversary wishes to associate the pushed item with every user in the dataset.
% In this attack the adversary does not distinguish between users, and wants to increase the overall frequency of the target item is recommended.
As this threat model does not assume the adversary has any knowledge of usage behavior, preferences of mass population need to be approximated. To this end, an adversary might pick $N$ random items, naturally without knowing which users have interacted with them. The assumption is if $N$ is large enough, then these items span diverse preferences.
% Then, injecting to the system a single user with a random list of items, would result with a noisy representation, which does not
Then the adversary adds these items by injecting $N$ users, each with a single item in their interaction history, to function as the set of target users. The \emph{active target} user is chosen in a round-robin manner, and an adversarial step is made to maximize the score for that user.

However, we noticed that this method fails, as maximizing the score for the active target user also deteriorates the scores of the pushed item for many other target users. This is an expected behavior, as it is infeasible to find a direction that increases the scores of an entire set of randomly selected users.

Keeping in mind that the objective of the adversary is not to increase the average score of the pushed item, but to bubble it up to the \emph{top} of many users' lists, we modify this approach by dynamically choosing the next active user. Preceding to each step, the adversary ranks the target users by their personalized score of the pushed item\footnote{Note that this operation adds $N$ parallelable queries to the recommender and does not require to alter the image.}. Then it concentrates on target users who are more likely to receive the pushed item in the top of their recommendation list. This is done by setting the active target user as the one in the $p^{th}$ percentile. As $p$ increases from 0 to $100\%$, the adversary concentrates on a larger portion of the target users. we tuned $p$ on 5 equally distanced values in the range $[0,1]$ and obtained optimal performance at $p=25\%$.

In this type of experiment we pick at random 100 pairs of items, each of which serves as the pushed item and set $N=200$ as the number of random items. Once again, the adversary cannot know the actual effectiveness of this attack on the general population, since their identities and usage patterns are not disclosed.
We report the performance of the attack on $1,000$ randomly chosen users who serve as a representative sample of the general population.

\subsection{Baselines}
Although the main purpose of this paper is to empirically demonstrate the existence of a threat on AI safety posed by visual attacks, it is still important to compare the effectiveness of our %finely designed
methods with relevant baselines.
To empirically demonstrate the existence of the identified vulnerability and to judge the relative effectiveness of the proposed attack models, we include two baseline attacks in our experiments. Since, to the best of our knowledge, this is the first work to investigate these types of visual attacks in the context of recommenders, there are no prior baselines.

Nevertheless, to show the importance of carefully designed perturbations, we propose two simple baseline methods that perform push attacks by altering genuine images.
\begin{itemize}
\item The first is an easy-to-detect attack, which replaces the original image $p$ of the pushed item with the image $p_{pop}$, associated with the most popular item from the same category. The rationale behind this baseline is that some of the success of the most popular items is attributed to their images. A visually-aware recommender should capture the visual features that correlate with consumption, and reflect it in superior scores for items with these images.
\item The second baseline is a softer version of the former, making it harder to be detected. In this baseline, $p\leftarrow p+\epsilon\cdot p_{pop}$, where $\epsilon$ controls the amount of change added to the image. We experimented values in the range $[0.01, 0.1]$ in steps of $0.01$.

\end{itemize}

However, while in all configurations both baselines systematically increased the scores of the pushed items and improved their ranking in thousands of places on average, they failed to penetrate to the top of the lists and to improve $HR@K$. The resulting Hit Ratios were therefore consistently very close to zero in all experiments, which is why we omit the results in the following sections. Ultimately,
this evidence reinforces our preliminary assumption on the importance of well-guided adversarial examples.

\subsection{Results}
Here, we report the results of the experiments and present multiple trade-offs between adversarial resources and effectiveness. Across all experiments, when partial ranking knowledge is available, we assume that only the top 1\% of the ranking is known.
% Since the main purpose of our paper is to show the potential of visual attacks, we focus on situations where a complete catalog ranking exists. However, we also report selected results when only \emph{partial} ranking knowledge is available. Specifically, in these situations we assume that only the top 1\% of the ranking is known.
%In our analyses, we mainly focus on the specific-user attack and on
% DJHERE
%\blue{We consider both situations where the complete ranking is known, and scenarios where only partial ranking knowledge is available. In the latter case, we assume that only 1\% of the ranking is revealed to the adversary (denoted as ``Partial'' in the figures).}
%Since the main purpose of our paper is to show the potential of visual attacks, we furthermore focus on situations where a complete catalog ranking exists.
%Because of the relative complexity of the attack under partial ranked list

\subsubsection{Specific-user Attack}
Table \ref{tab:specific_users} details the HR$@k$ of different attacks for various values of $k$. We fixed the number of perturbations (32) and steps (20), leading to a total of $640=32\cdot 20$ image updates.
Attack types denoted by WB, BB-Score, BB-Rank and BB-Partial stand for white-box, black-box on scores, black-box on rankings and black-box on partial ranking knowledge, respectively.

The results demonstrate the effectiveness of all proposed attack models. In terms of the absolute numbers for the Hit Ratio we can observe that after the attack the probability of a randomly pushed item to appear in the top-20 list is substantial, even when using a black-box approach. Generally, the results also validate the existence of trade-off between knowledge on the attacked system and effectiveness. We notice that the white-box attack, which assumes absolute knowledge on the underlying system, as expected outperforms the rest of the attacks. However, the differences between the black-box approaches on rankings and scores are usually small and depend on the metric and dataset.

%There is a negligible advantage to score-feedback over ranking only. The minor difference between them proves the ability of our proposed approach to estimate scores.
% Therefore, throughout the rest of the experiments, include only
It is worthwhile to mention that even in the Electronics dataset, where visual features play a minor role, with attributed gain of only $2\%$ to the AUC, the attack is still effective. For a considerable amount of users, it can bubble up random items to the top of their lists. This proves that under attack, even seemingly less important side-information can be used to manipulate the results of collaborative filtering models.

\begin{table}%[H]
\begin{center}
\begin{tabular}{ |m{4.2em}|m{3.8em}|c||m{2.3em}|m{2.7em}|m{2.7em}| }
\hline
Dataset & RS & Attack & $HR@1$ & $HR@10$ & $HR@20$ \\
\hline
\multirow{8}{*}{Clothing} &
\multirow{4}{*}{VBPR} & WB & 0.79 & 0.89 &0.91 \\
\cline{3-6}
& & BB-Score & 0.43 & 0.54 & 0.58 \\
\cline{3-6}
& & BB-Rank & 0.41 & 0.51 & 0.53 \\
\cline{3-6}
& & BB-Partial & 0.23 & 0.23 & 0.23 \\
\cline{2-6}
& \multirow{4}{*}{DeepStyle} & WB & 0.73 & 0.80 &0.83 \\
\cline{3-6}
& & BB-Score & 0.39 & 0.50 & 0.52 \\
\cline{3-6}
& & BB-Rank & 0.42 & 0.50 & 0.54 \\
\cline{3-6}
& & BB-Partial & 0.10 & 0.10 & 0.10 \\
\hline
\multirow{8}{*}{Electronics} &
\multirow{4}{*}{VBPR} & WB & 0.45 & 0.57 & 0.60 \\
\cline{3-6}
& & BB-Score & 0.15 & 0.22 & 0.27 \\
\cline{3-6}
& & BB-Rank & 0.14 & 0.24 & 0.27 \\
\cline{3-6}
& & BB-Partial & 0.47 & 0.50 & 0.50 \\
\cline{2-6}
& \multirow{4}{*}{DeepStyle} & WB & 0.53 & 0.62 & 0.66 \\
\cline{3-6}
& & BB-Score & 0.18 & 0.20 & 0.29 \\
\cline{3-6}
& & BB-Rank & 0.20 & 0.26 & 0.29 \\
\cline{3-6}
& & BB-Partial & 0.20 & 0.20 & 0.20 \\
\hline
\end{tabular}
\end{center}
\caption{Efficacy of specific-user attack}
\label{tab:specific_users}
%\vspace{-0.4cm}
\end{table}

Due to space limitations, in the rest of the paper we report the results only of a subset of the configurations.
First, as we did not find any significant difference in terms of the attack effects for the  different algorithms (VBPR and DeepStyle) we focus on the former. Second, while the HR of Clothing and Electronics are different, their trends across the experiments remain the same. Hence, we report the results of the Clothing dataset. Third, the default value of $k$ is 20 in the $HR@k$ measure.
Moreover, in the experimented datasets across the three color channels, pixels take values in the range $(0,255)$. We report the results obtained by setting the minimal step size $\epsilon$, which bounds the size of each update to at most $1$, as it led to superior performance comparing to higher values.

\begin{figure}[h!t]
%\begin{minipage}[b]{0.485\linewidth}
\includegraphics[trim={1.3cm 0.4cm 2cm 1cm}, clip, scale=0.41, center]{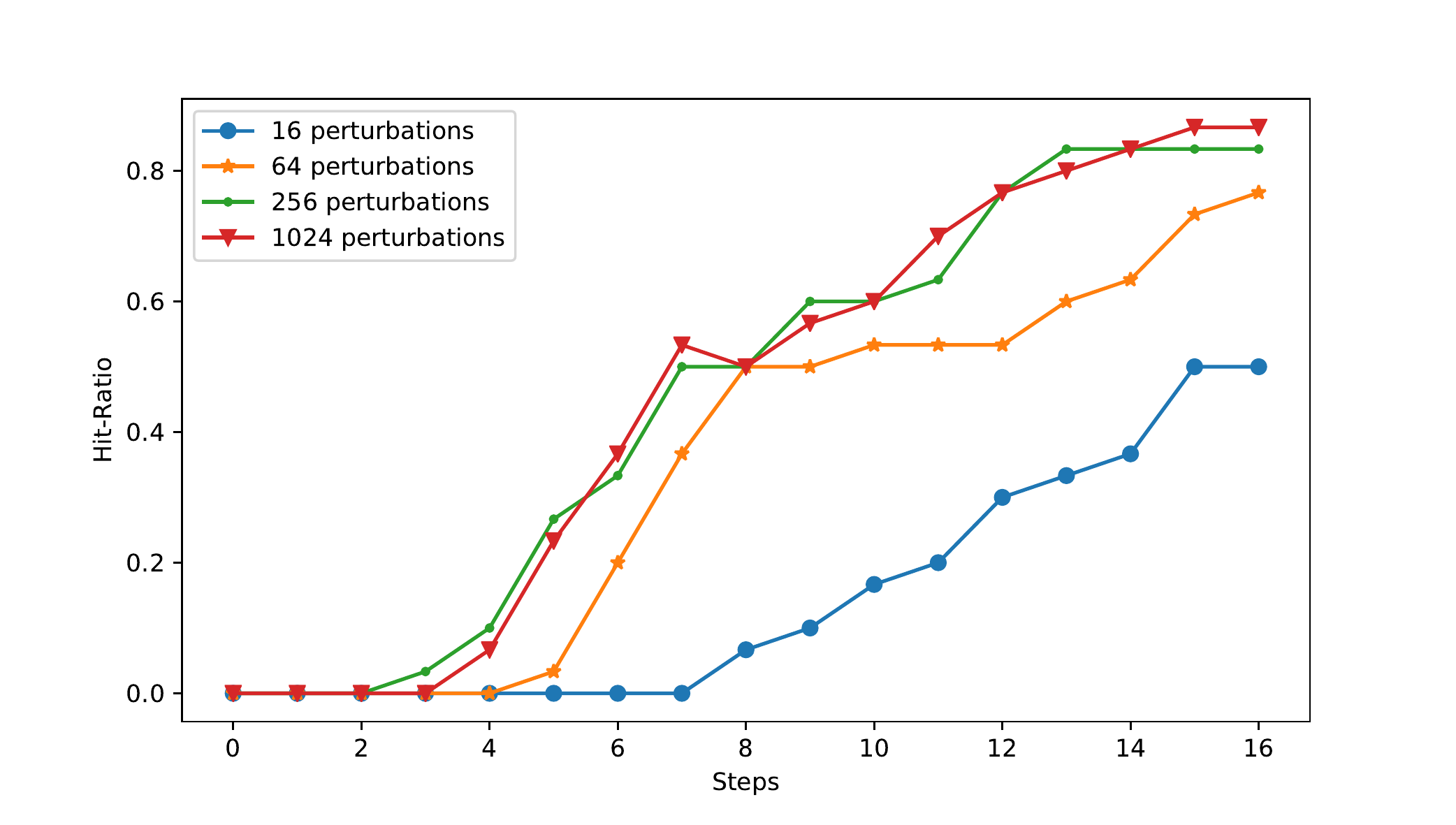}
\caption{Effect of number of perturbations and update steps}
% \vspace{-0.9cm}
\label{fig:perturbations}
%\end{minipage}
%\quad
\end{figure}

\begin{figure}[h!t]
%\begin{minipage}[b]{0.485\linewidth}
\includegraphics[trim={1.3cm 0.4cm 2.cm 1cm}, clip, scale=0.41, center]{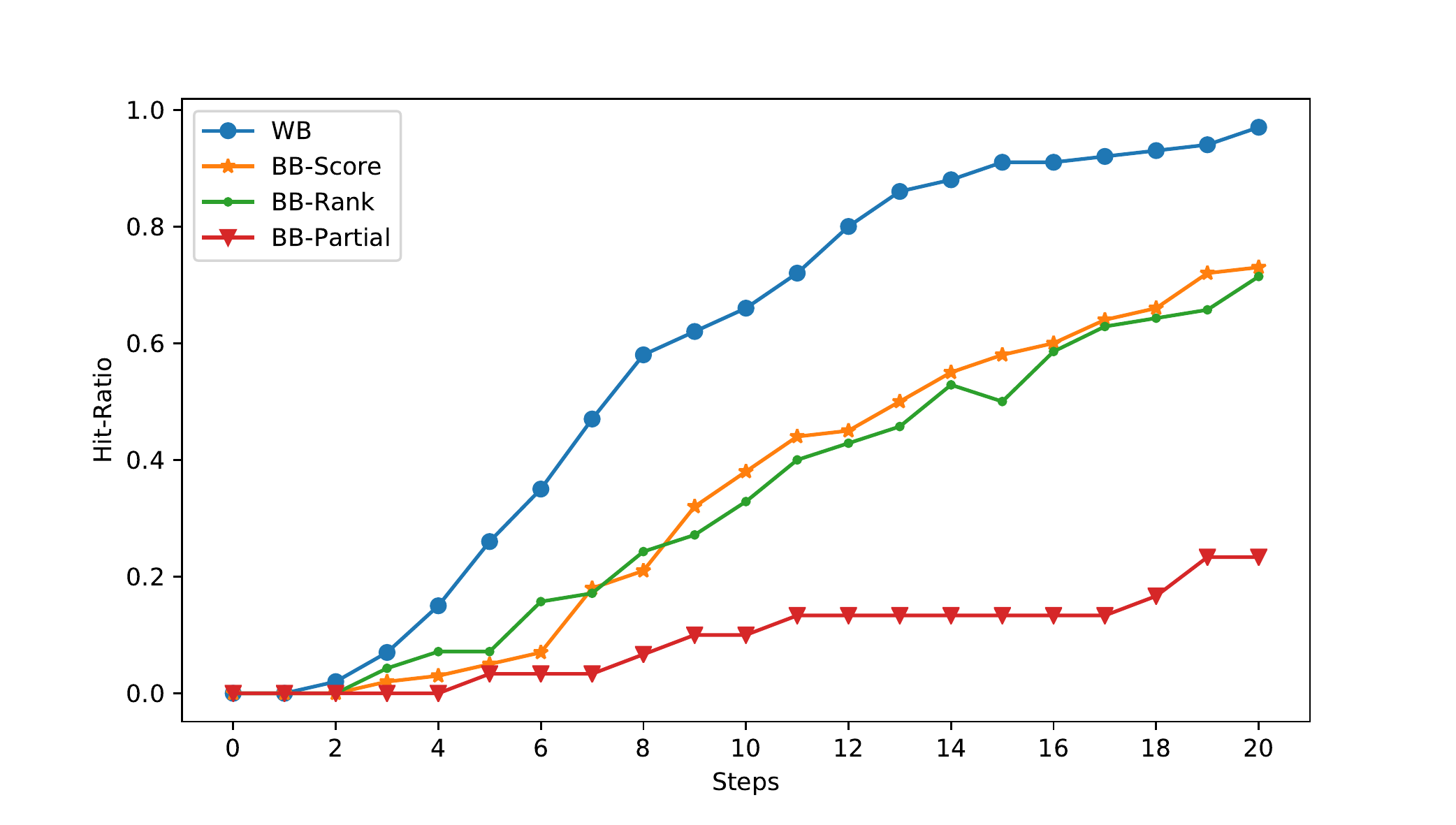}
\caption{Effectiveness of each attack over number of steps}
\label{fig:steps}
%\end{minipage}
\end{figure}

Figures \ref{fig:perturbations} and \ref{fig:steps} extend Table \ref{tab:specific_users} and probe the effects of isolated factors.
Figure \ref{fig:perturbations} visualizes the trade-offs between effectiveness, number of steps, and number of perturbations under the BB-Rank attack. It plots HR as a function of the number of steps, for several values of perturbations per step.

As can be seen, if the adversary employs 64 perturbations per step and wants to obtain HR of roughly 0.5, then 8 steps are required, which means a total of $512=64\cdot 8$ image updates. In comparison, obtaining on-par performance with 16 perturbations requires to almost double the number of steps, and to take 15 steps, which might deteriorate the visualization of the image. However, that alternative requires only $240=16\cdot 15$ image updates.
We note that using 2,048 perturbations leads to a single solution of the gradient, without ability to generalize. Its effect is detrimental and these attacks fail to increase the HR.

In the rest of the experiments we fix the number of perturbations to 64, as it achieves solid performance with a moderate budget of image updates. Figure \ref{fig:steps} shows the effectiveness of each attack at various numbers of steps.
We first observe that the effectiveness of the four types of attacks monotonically increases with the number of steps. This demonstrates the effectiveness of our computations,
as each additional step directs the adversary to achieve more impact.
% It is clear that the white-box attack converges much faster than the other two black-box attacks.
The BB-Score and BB-Rank behave very similarly throughout the attack, which shows our proposed method of estimating scores by ranking is consistent.
Ranking based on partial knowledge also leads to solid performance, which shows the potential of this attack in real-life scenarios.

% \begin{figure}[H]
% \includegraphics[trim={1.3cm 0.4cm 2cm 1cm}, clip, scale=0.47]{steps.pdf}
% \caption{Efficiency of each attack over steps}
% \label{fig:steps}
% \end{figure}

% \begin{figure}[hbt!]
% \includegraphics[trim={1.3cm 0.4cm 2cm 1cm}, clip, scale=0.47]{perturbations.pdf}
% \caption{Effect of number of perturbations and update steps}
% \label{fig:perturbations}
% \end{figure}

\begin{figure}[h!t]
%\begin{minipage}[b]{0.485\linewidth}
\includegraphics[trim={1.37cm 0.4cm 2.1cm 1cm}, clip, scale=0.41, center]{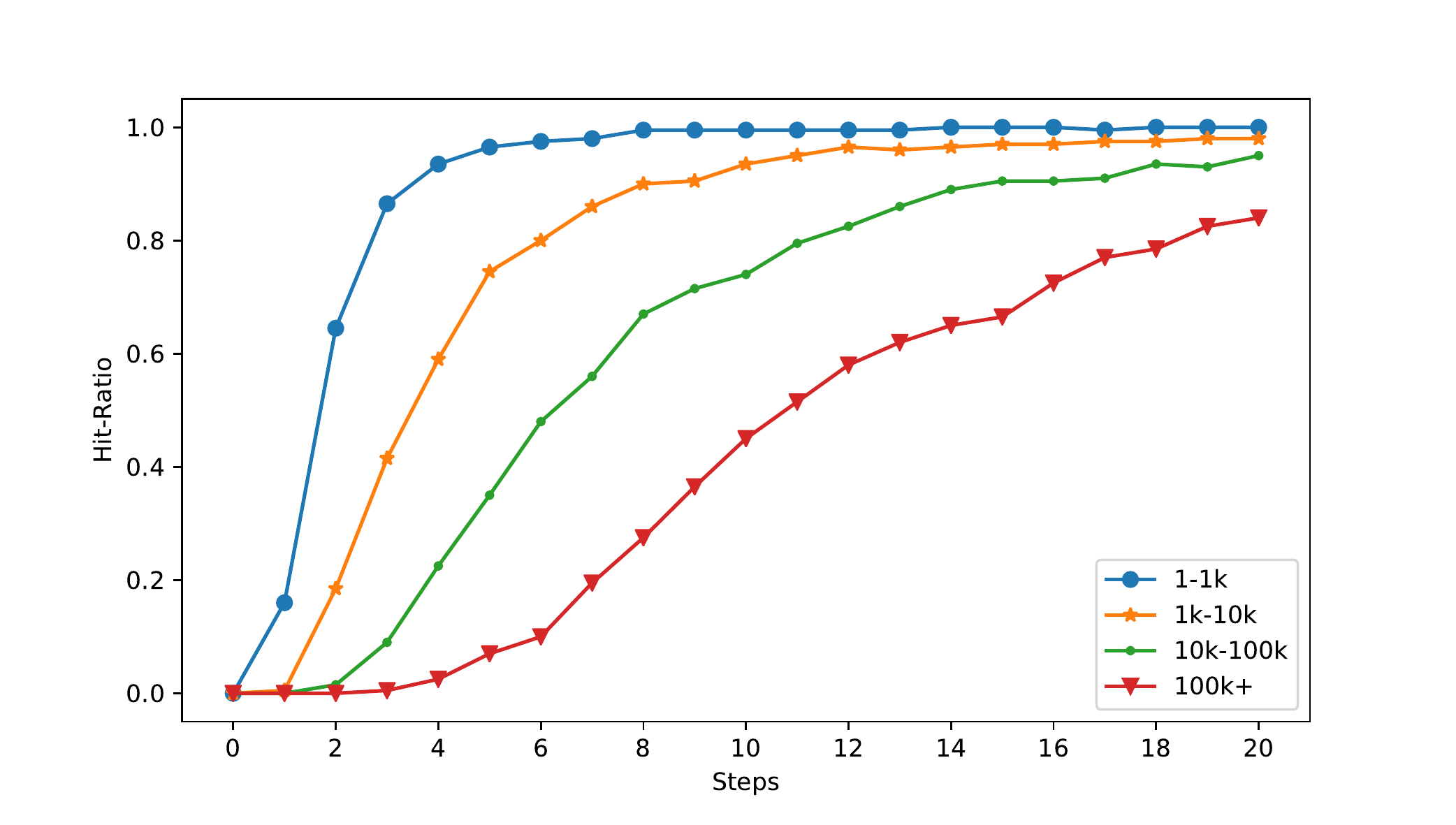}
\caption{Effectiveness over different rank ranges}
\label{fig:from_rank}
\end{figure}

Finally, we hypothesize that the effectiveness of an attack depends on the relevance of the pushed item to the target user, approximated by the pre-attack ranking. Figure \ref{fig:from_rank} shows the effectiveness of the BB-Rank attack where the pushed items are grouped by their initial ranking. As it shows, attacks on relevant items converge faster and achieve higher HR. Remarkably, even when the pushed item is irrelevant to the user, i.e., the initial ranking is in the hundreds of thousands, within a plausible number of 20 steps, the adversary succeeds in pushing the item to the top of the lists of the vast majority of the target users.
% The level of noise controls how effective the attacks can be without making too many steps which will make the image too noisy. Figure \ref{fig:from_rank} shows the effectiveness of the black-box attack where the pushed items group. It depicts the convergence rate of to make items from different initial rank-ranges to reach the top of the list. Even for items that ranked initially at rank 100,000 and above, the HR@20 is over 0.4 after 10 steps and over 0.8 after 20 steps.

%\end{minipage}
%\quad
%\begin{minipage}[b]{0.485\linewidth}

% \begin{figure}
% \includegraphics[trim={1.3cm 0.4cm 2cm 1cm}, clip, scale=0.47]{from-rank.pdf}
% \caption{Efficiency over different rank ranges}
% \label{fig:from_rank}
% \end{figure}

\subsubsection{Segmented and General Population Attacks}
\begin{figure}[h!t]
% \vspace{-0.3cm}
\includegraphics[trim={0.5cm 0.4cm 2cm 1cm}, clip, scale=0.41, center]{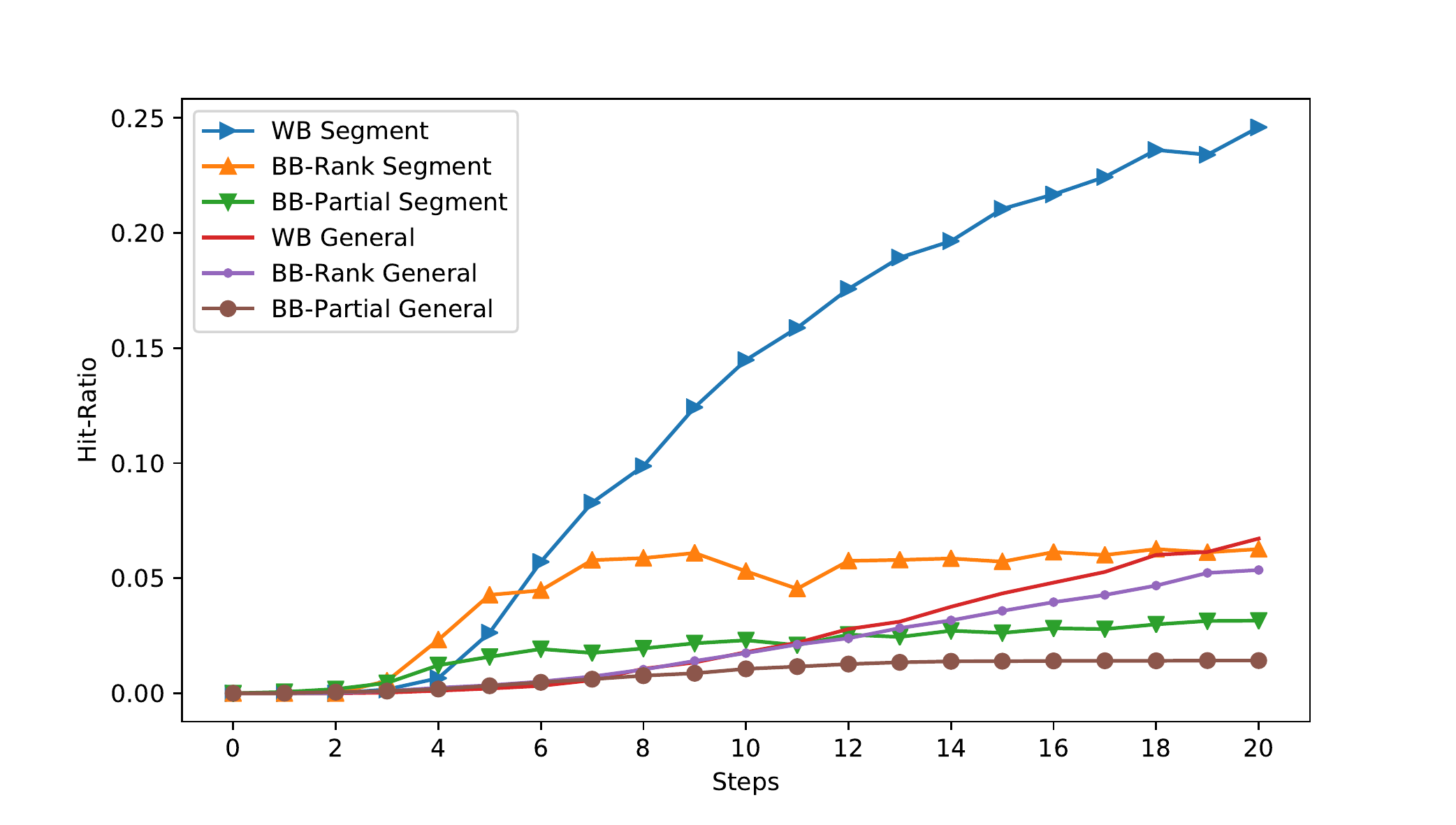}
\caption{Segmented and general population attacks}
\label{fig:segment_general}
%\end{minipage}
% \vspace{-0.5cm}
\end{figure}

Figure \ref{fig:segment_general} shows the performance of the segmented and general population attacks side by side. Again, we can observe a correlation between the adversary's knowledge about the target users and performance. Compared to the general population attack, the Hit Ratio is higher for the segmented attack, where the adversary knows a single item in the history of all users (although does not know the identity of them). Furthermore, both of these attacks are outperformed by the specific-user attacks (not shown in the figure), in which the adversary possesses vast knowledge on the interaction history of the users. This comes as no surprise, since deep knowledge on users entails more accurate guidance for the adversary.
When given only \emph{partial} ranking knowledge, the attack is still successful for a significant amount of users, e.g., around 1.6\% of the entire user base for the general population attack.
%.\footnote{Remember that the reported values relate to percentages of the entire user base.}}
%Additionally, we can observe that even when the top 1\% percent of the items are available, the attack can still succeed for a significant amount of users.

% We note that even though the HR of the general population attack under the BB-Rank model reaches only few percents, it reflects a high volume appearances in the top lists, because it approaches to mass population.

% \begin{figure}
% \includegraphics[trim={0.5cm 0.4cm 2cm 1cm}, clip, scale=0.45]{segment-general-per25.pdf}
% \caption{Segmented and general population attacks}
% \label{fig:segment_general}
% \end{figure}

\section{Discussion and Outlook}\label{sec:discussion}
In this paper we identified and investigated a new type of vulnerability of recommender systems that results from the use of externally-provided images. We devised different white-box and, more importantly, black-box attack models to exploit the vulnerabilities. Experiments on two datasets and showed that the proposed attack models are effective in terms of manipulating the scores or rankings of two visually-aware RS.

%therefore they present a new and concrete weakness in AI safety.

While the proposed approaches rely on images to perform attacks, the same techniques can be applied on any continuous side information. Consequently, a possible future work is to investigate the vulnerabilities of audio-based \cite{van2013deep}, video-based \cite{covington2016deep} or textual review-based \cite{shalom2019generative} recommenders .
% adversarial examples are a good aspect of security to work on because they represent a concrete problem in AI safety that can be addressed in the short term, and because fixing them is difficult enough that it requires a serious research effort.

% discussion - if the attack is too efficient, will be detected

% Now, after demonstrating the clout of various attack types, it is of great importance

We hope this paper will raise awareness about image-based attacks and spur research on means to combat such attacks.
%this detrimental behavior by exploiting the vulnerabilities of the attacker.
In the context of defending RS against shilling attacks, we are encouraged by the impressive effectiveness of detection algorithms \cite{burke2006classification,williams2007defending,li2016shilling,zhang2018ud,tong2018shilling}, and the results of robust RS \cite{turk2018robust,alonso2019robust}. Also, existing works on increasing the robustness of RS against adversarial examples \cite{he2018adversarial,du2018enhancing} may ignite research on robustness against visual-attacks. Additionally, research against adversarial examples in general is fruitful and several algorithms were proposed for detection \cite{carlini2017adversarial,xu2017feature} and new robust models were devised \cite{burke2006classification,gu2014towards}.
We anticipate a plethora of work in these veins towards more robust recommenders, which
% , e.g., simply limit the number of allowed image updates or
are more robust to tiny visual changes.

%In fact, there could be diverse plausible approaches to mitigate this threat, ranging from simply limiting the number of allowed image updates until applying robust algorithms that do not alter their recommendations due to tiny visual changes.
% One plausible approach emanates from the need of the attacker for multiple perturbations of original image. The recommender system's admin may log those changes, and act upon them. For instance, it may be worthwhile to limit the number of allowed changes of each item, or to set a minimal distance between images.
\newpage
\bibliographystyle{ACM-Reference-Format}
\bibliography{sample-sigconf}

\end{document}